\UseRawInputEncoding

\documentclass[preprint,12pt,authoryear]{elsarticle}

\usepackage{amssymb}
\usepackage{lineno}
\usepackage{caption}
\usepackage{color}
\usepackage{hyperref}
\hypersetup{colorlinks=true}
\usepackage{multirow}
\usepackage[normalem]{ulem}
\useunder{\uline}{\ul}{}
\usepackage{subfigure}
\usepackage{algorithm}
\usepackage{algorithmic}
\usepackage[capitalize]{cleveref}
\usepackage{tabularx}

\journal{Neural Network}

\begin{document}
\begin{frontmatter}

\title{A Dual-Branch Model with Inter- and Intra-branch Contrastive Loss for Long-tailed Recognition}\pdfoutput=1 %For ARXIV

\author[1]{Qiong Chen\corref{cor1}}\pdfoutput=1 %For ARXIV
\ead{csqchen@scut.edu.cn}\pdfoutput=1 %For ARXIV
\author[1,2]{Tianlin Huang}\pdfoutput=1 %For ARXIV
\ead{htanry@126.com}\pdfoutput=1 %For ARXIV
\author[1]{Geren Zhu}
\ead{gr.zhu@aliyun.com}
\author[1]{Enlu Lin}
\ead{linenus@outlook.com}
\cortext[cor1]{Corresponding author at: School of Computer Science and Engineering, South China University of Technology}
\affiliation[1]{organization={School of Computer Science and Engineering, South China University of Technology},%Department and Organization
            % city={Guangzhou},
            % postcode={510006}, 
            country={China}}
\affiliation[2]{organization={Guangdong Provincial Key Laboratory of Artificial Intelligence in Medical Image Analysis and Application},
            country={China}}

\begin{abstract}
Real-world data often exhibits a long-tailed distribution, in which head classes occupy most of the data, while tail classes only have very few samples. Models trained on long-tailed datasets have poor adaptability to tail classes and the decision boundaries are ambiguous. Therefore, in this paper, we propose a simple yet effective model, named Dual-Branch Long-Tailed Recognition (DB-LTR), which includes an imbalanced learning branch and a Contrastive Learning Branch (CoLB). The imbalanced learning branch, which consists of a shared backbone and a linear classifier, leverages common imbalanced learning approaches to tackle the data imbalance issue. In CoLB, we learn a prototype for each tail class, and calculate an inter-branch contrastive loss, an intra-branch contrastive loss and a metric loss. CoLB can improve the capability of the model in adapting to tail classes and assist the imbalanced learning branch to learn a well-represented feature space and discriminative decision boundary. Extensive experiments on three long-tailed benchmark datasets, i.e., CIFAR100-LT, ImageNet-LT and Places-LT, show that our DB-LTR is competitive and superior to the comparative methods.
\end{abstract}

% %Graphical abstract
% \includegraphics[width=0.95\textwidth, height=0.5\textwidth]{DC-LTR-revised.pdf}

% %Research highlights
% \begin{highlights}
% \item A dual-branch long-tailed recognition model is proposed to improve the adaptability of the model to tail classes and the distinguishability of decision boundary.
% \item An inter-branch contrastive loss is designed to distinguish head- and tail features.
% \item An intra-branch contrastive loss is designed to make the class boundary of tail classes distinct.
% \item A prototype-based metric loss is designed to boost the model's adaptation ability to tail classes.
% \item Comprehensive experiments show that our proposed method is superior to the comparative methods.
% \end{highlights}

\begin{keyword}
Neural network \sep long-tailed recognition \sep imbalanced learning \sep contrastive learning


\end{keyword}

\end{frontmatter}

% \linenumbers
% \pagewiselinenumbers

\section{Introduction}
\label{introduction}
In recent years, with various large-scale and high-quality image datasets being easily accessed, deep Convolutional Neural Networks(CNNs) have achieved great success in visual recognition \citep{he2016deep, zhang2018cross, liang2020paying}. These datasets are usually artificially balanced, however, real-world data often exhibits a long-tailed distribution, where a few classes (majority or head classes) occupy most of the data, while other classes (minority or tail classes ) only have very few samples. Due to the extreme class imbalance and limited tail samples of the long-tailed datasets, the feature representation of tail classes learned by the model is insufficient. Consequently, the model has poor adaptability to tail classes and the decision boundaries are ambiguous.

Several kinds of methods have been proposed to improve the classification of  tail classes, including class re-balancing, data augmentation and ensemble model. Class re-balancing can be roughly divided into re-sampling \citep{buda2018systematic, pouyanfar2018dynamic, chawla2002smote, drummond2003c4} and re-weighting \citep{wang2017learning, park2021influence, shu2019meta}, both methods share the goal of rectifying the skewed training data to improve the classifier learning. Data augmentation strategy \citep{zhang2017mixup, chou2020remix, li2021metasaug, wang2021rsg} generates new information for tail classes to improve the recognition capability of tail samples. Ensemble model \citep{cai2021ace, wang2020long, cui2022reslt, xiang2020learning} splits the long-tailed datasets into several relatively balanced subsets and trains a model with multiple classifier, embodying the design philosophy of divide-and-conquer. These approaches reshape the decision boundary and improve the generalization ability of the model via enhancing the learning law on tail classes.

However, the class re-balancing either takes the risk of overfitting to tail classes and under-learning of head classes\citep{wang2021contrastive, yang2022survey}. The data augmentation strategy lacks theoretical guidance and cannot introduce new efficient samples \citep{yang2022survey}. And the powerful computational resources underpin the great success of ensemble model.

In this paper, we improve the long-tailed model’s adaptation ability to tail classes and the decision boundaries with a Dual-Branch Long-Tailed Recognition (DB-LTR) model, which is built upon an imbalanced learning branch and a contrastive learning branch. The imbalanced learning branch aims to cope with the data imbalance and model bias. In Contrastive Learning Branch (CoLB), tail samples are used to supervise the model training under the $N$-way $N_{sup}$-shot setting \citep{snell2017prototypical}. CoLB only utilizes a small amount of tail data, which avoids overfitting and does not bring much more computational overhead for our method.

In DB-LTR, the imbalanced learning branch performs common imbalanced learning methods, e.g., LDAM \citep{cao2019learning}, Remix \citep{chou2020remix}, etc. The key idea underlying this branch is to train an unbiased or less biased classifier. Nevertheless, the issue of insufficient feature learning of tail classes remains untouched. Therefore, we introduce CoLB to tackle this problem. To be specific, we calculate a prototype-based \citep{snell2017prototypical} metric loss to promote the model’s adaptation to tail classes. In order to separate head- and tail features and to make the class boundaries of tail classes distinct in embedding space, we borrow contrastive learning to calculate an inter-branch Contrastive Loss (inter-CL) and an intra-branch Contrastive Loss (intra-CL), respectively. Benefiting from these two contrastive losses, CoLB can learn a decision boundary that keeps far away from head- and tail classes simultaneously. We have validated the effectiveness of DB-LTR on three long-tailed datasets.

The main contributions of this paper are summarized as follows:
\begin{enumerate}[(1)]
    \item We propose a Dual-Branch Long-Tailed Recognition model, which is composed of an imbalanced learning branch and a contrastive learning branch. Our model can effectively enhance the distinguishability of decision boundary and the adaptability of the model to tail classes.
    \item The Contrastive Learning Branch (CoLB) can seamlessly and friendly plug in other imbalanced learning methods to further improve their recognition performance.
    \item Experimental results on three long-tailed benchmarks, i.e., CIFAR100-LT, ImageNet-LT and Place-LT, show that our method poses promising performance and surpasses the comparative methods.
\end{enumerate}

\section{Related work}
\label{related_work}
\subsection{Long-tailed recognition}
Existing methods of addressing long-tailed problems include class re-balancing, data augmentation, decoupled learning and ensemble model \citep{cai2021ace, wang2020long, cui2022reslt, xiang2020learning}.

\textbf{Class re-balancing.} Traditional class re-balancing methods are mainly divided into re-sampling and re-weighting. Re-sampling reduces the overwhelming effect of head class on the model during training via under-sampling head classes \citep{drummond2003c4, buda2018systematic} or over-sampling tail classes \citep{buda2018systematic, chawla2002smote, more2016survey}. Some recent works attempt to explore more effective sampling strategies \citep{pouyanfar2018dynamic, kang2020decoupling}. Re-weighting assigns different loss contribution \citep{lin2017focal, wang2017learning} or decision margin \citep{cao2019learning} to different classes according to the frequencies or difficulties of samples. CB Loss \citep{cui2019class} decides the weighting coefficients based on the effective number of samples. These approaches adjust the decision boundary in an indirect way. Different from them, \citet{park2021influence} re-weight each sample according to their influence on the decision surface, aiming to learn a generalizable decision boundary. Since class re-balancing methods enlarge the importance of tail classes during model optimization, they can effectively promote the generalization and classification of the under-performed minority classes.

\textbf{Data augmentation.} Unlike re-sampling, data augmentation strategy \citep{chou2020remix, wang2021rsg} aims to exterminate the ambiguity of decision boundary via generating new additional information for tail classes. Remix \citep{chou2020remix} decomposes the interpolation of samples and labels, which enlarges the importance of tail classes when interpolating labels, while the samples interpolation employs standard mixup \citep{zhang2017mixup}. \citet{wang2021rsg} observe that the scarcity of tail samples impels the learned feature space of the rare classes to be too small. Hence, they design a generator dedicated to augmenting tail features so that alleviate the misclassification of tail data. ISDA \citep{wang2019implicit} brings inferior performance in long-tailed scenarios because of limited samples in tail classes. MetaSAug \citep{li2021metasaug} overcomes this shortcoming. It learns transformed semantic directions in the way of meta-learning to perform meaningful semantic augmentation for tail classes, resulting in the improvement of classifier learning.

\textbf{Decoupled learning.} Decoupled learning \citep{kang2020decoupling, wang2022dynamic} optimizes the feature backbone and classifier using different training strategies. Recent works \citep{zhang2021distribution, zhong2021improving} aim to make the decision surfaces more recognizable by retraining the classifier. \citet{zhang2021distribution} observe that the performance bottleneck of the long-tailed model is the biased decision boundary. Thus, they propose an adaptive calibration function and a distribution alignment strategy to re-balance the classifier. The same viewpoint of decoupling can be found in \citep{alshammari2022long}, which balances the classifier weight norm by tuning the weight decay properly. The ambiguity of decision boundary implies bias residing in classifier, which leads to miscalibration and over-confidence. \citet{zhong2021improving} tackle these issues via label-aware smoothing and shifted batch normalization.

\subsection{Contrastive learning}
Class re-balancing will lead to sub-optimal feature representation \citep{kang2020decoupling, zhou2020bbn, wang2022dynamic}. On the contrary, contrastive learning is a promising work in unsupervised representation learning \citep{chen2020simple, he2020momentum, kang2020exploring}. There is the insight that contrastive learning learns a compact and separable feature space by pulling together the samples belonging to the same classes while pushing away the samples of different categories. Recently, Supervised Contrastive Learning (SCL) poses considerable success in long-tailed recognition \citep{wang2021contrastive, cui2021parametric, li2022targeted}. The basic idea is that better features make better classifier. So \citet{wang2021contrastive} introduce a supervised contrastive loss to improve image representation and use the cross-entropy loss to train an unbiased classifier. \citet{kang2020exploring} claim that the feature space learned by self-supervised contrastive learning is more balanced than that of SCL yet lacks semantic discriminativeness. So they devise KCL \citep{kang2020exploring} to address this problem. The proposed Hybrid-PSC \citep{wang2021contrastive} and BCL \citep{zhu2022balanced} are the most relevant to our method. Hybrid-PSC contrasts each sample against the prototypes of all other categories, and BCL aims to learn an ideal geometry for representation learning by class-averaging and class-complement. Although similar, our DB-LTR differs from Hybrid-PSC and BCL as we introduce CoLB dedicated to dealing with the poor adaptability of the model to tail classes and the ambiguity of decision boundary.

\section{Methodology}
\label{methodology}
\subsection{Overall framework}
\begin{figure*}[!h]
    \centering
    \includegraphics[width=0.95\textwidth, height=0.5\textwidth]{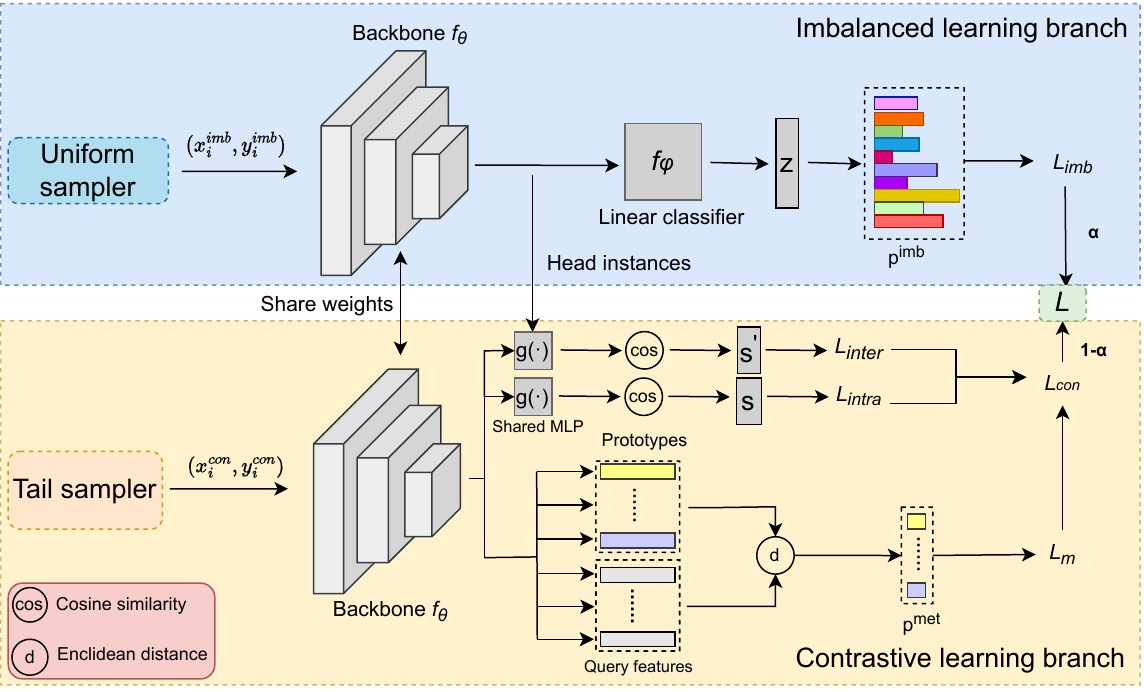}
    \caption{The architecture of our Dual-Branch Long-Tailed Recognition (DB-LTR) model. The uniform sampler utilizes instance-balanced sampling on the whole long-tailed datasets, while the tail sampler only samples tail data randomly. The head instances from the imbalanced learning branch are used to compute the inter-CL ($L_{inter}$). For intra-CL ($L_{intra}$), it contrasts the support- and query samples inside the contrastive learning branch. The prototype-based metric loss ($L_m$) is calculated using non-parametric metric.}
    \label{fig:architecture}
\end{figure*}

The architecture of our DB-LTR is shown in Figure \ref{fig:architecture}, it mainly consists of an imbalanced learning branch and a contrastive learning branch. Therein, the imbalanced learning branch includes a shared backbone $f_{\theta}$ and a linear classifier $f_{\varphi}$. This channel acts as the main branch and integrates common imbalanced learning methods, e.g., LDAM \citep{cao2019learning}, Remix \citep{chou2020remix}, etc. Contrastive Learning Branch (CoLB) is an auxiliary branch, which is dedicated to enhancing the model’s adaptation ability and to learning a compact and separable feature space. Specifically, to compensate the under-learning of tail classes, CoLB samples tail data to supervise the model learning through calculating an inter-branch contrastive loss (inter-CL), an  intra-branch contrastive loss (intra-CL) and a metric loss. The two branches share the same backbone structure and weights. In the training phase, the total loss is defined as the weighted loss of these two branches and is used to update the network. During the inference phase, the shared backbone and the linear classifier are used to make predictions.

\subsection{Imbalanced learning branch}
The imbalanced learning branch is used to mitigate the influence of data imbalance and the model’s preference for head classes. We integrate LDAM \citep{cao2019learning} into this branch, which encourages model to learn different margins for different classes. That is, if a class has a large number of samples, it will be assigned a relatively small margin, otherwise, a large margin will be allocated.

Let $z_i=f_\varphi (f_\theta (x_i^{imb}))$ denote the logit for a training sample $x_i^{imb}$. According to LDAM, the probability of sample $x_i^{imb}$ to be predicted as class $j$ is:
\begin{equation}
\label{equation1}
    p_{i,j}^{imb}=\frac{\exp\left ( z_j-\Delta _j\right )}{\exp\left ( z_j-\Delta _j\right )+\sum_{i\neq j}\exp\left ( z_i\right )}
\end{equation}
where $\Delta _j=H/ \left ( n_j^{1/4}\right )$ denotes the margin for class $j$, $n_j$ is the number of samples in $j$-th class, $H$ is an adjustable hyper-parameter. The more samples there are in a class, the smaller margin the class obtains.

Cross-entropy loss is used to calculate the loss of the imbalanced learning branch:
\begin{equation}
\label{equation2}
    L_{imb}=-\frac{1}{N_{imb}}\sum _{i=1}^{N_{imb}}L_{CE}\left ( p_{i}^{imb}, y_i^{imb}\right )
\end{equation}
where $N_{imb}$ is the number of samples in the imbalance learning branch, $y_i^{imb}$ is the label of sample $x_i^{imb}$.

Other imbalanced learning methods can also be integrated into this branch. When collaborating with CoLB, they achieve better performance. More details are described in Section \ref{ablation_study}.

\subsection{Contrastive learning branch}
In order to make the model better adapt to tail classes and learn generic features, a Contrastive Learning Branch (CoLB) is designed. In CoLB, we learn a prototype representation \citep{snell2017prototypical} for each tail class, and leverage tail samples to calculate a prototype-based metric loss, aiming to improve the model's ability to recognize tail classes. In long-tailed datasets, due to the small number of tail samples, they are overwhelmed by the dominant head instances, which leads to insufficient and sub-optimal feature learning. Therefore, we further calculate an inter-branch Contrastive Loss (inter-CL) and an intra-branch Contrastive Loss (intra-CL), aiming to learn a feature space with the property of intra-class compactness and inter-class separability.

To alleviate the over-suppression by head classes and to balance the model learning, the CoLB takes tail samples as inputs. To this end, we devise a tail sampler to sample training data for CoLB. In a mini-batch, firstly the tail sampler randomly selects $N$ categories on tail classes, and then randomly samples $\left (N_{sup}+N_{qry}\right )$ instances for each category, where $N_{sup}$ instances and $N_{qry}$ ones are used as support set and query set, respectively.

We denote the support set as $S=\left \{\left ( x_i^{sup},y_i^{sup}\right )\right \}_{i=1}^{N_{sup}}$ and the query set $Q=\left \{\left ( x_i^{qry},y_i^{qry}\right )\right \}_{i=1}^{N_{qry}}$, where $x_i$ represents a support/query sample and $y_i$ is its corresponding label. Without loss of generality, the support and query samples can be uniformly expressed as $\left ( x_i^{con},y_i^{con}\right )$. In CoLB, the samples are predicted via non-parametric metric. Specifically, we calculate the semantic similarity of samples with respect to the prototypes to obtain their prediction probability.

Let $S_j$ denote the set of samples in class $j$, and $c_j$ denote the prototype of class $j$. For any sample $x_i^{qry}$ in the query set, we can get its prediction probability by calculating the similarity of its feature and the prototype of class $j$:
\begin{equation}
\label{equation3}
    p_{i,j}^{met}=\frac{\exp\left ( -d\left ( f_{\theta}\left ( x_i^{qry}\right ),c_j\right )\right )}{\sum_k \exp\left ( -d\left ( f_\theta\left ( x_i^{qry}\right ),c_k\right )\right )}
\end{equation}
where $d$ is the Euclidean distance, $p_{i,j}^{met}$ can be interpreted as the probability that the sample $x_i^{qry}$ is predicted as class $j$.

We utilize cross-entropy loss to calculate the metric loss:
\begin{equation}
\label{equation4}
    L_m=-\frac{1}{N\times N_{qry}}\sum _{n =1}^{N}\sum _{i=1}^{N_{qry}}L_{CE}\left ( p_{i}^{met},y_i^{qry}\right )
\end{equation}

In order to make the class boundaries of tail classes distinguishable, intra-CL computes the contrastive loss between query samples and the prototypes of the support set in CoLB. Let $c'_j$ denote the prototype of support set $S_j$. For a query sample $x_i^{qry}$, supposing it belongs to class $j$. Therefore, the positive of $x_i^{qry}$ is the prototype $c'_j$, while except for category $j$, the prototypes of all classes are negatives. Then we calculate the cosine similarity $s_{i,j}$ between $x_i^{qry}$ and $c'_j$, as done in \citep{chen2020simple}. The contrastive loss of the positive pair $\left ( x_i^{qry}, c'_j\right )$ is defined as:

\begin{equation}
\label{equation5}
    s_{i,j}=\frac{g\left( f_{\theta} \left( x_{i}^{qry}\right)\right)\cdot c'_{j}}{\left \| g\left( f_{\theta} \left( x_{i}^{qry}\right)\right)\right \|\times \left \| c'_{j}\right \|}
\end{equation}
\begin{equation}
\label{equation6}
    l_{intra}\left ( x_i^{qry}, c'_j\right )=-\log\frac{\exp\left ( s_{i,j}/\tau\right )}{\sum_{n=1}^{N}\mathbb{I}\left ( n\neq j\right )\exp\left ( s_{i,n}/\tau\right )}
\end{equation}

where $g\left( \cdot \right)$ is a projection head and proven to be more appropriate for contrastive loss \citep{chen2020simple}, $\left \| \cdot \right \|$ denotes the L2 norm, $\mathbb{I}\left ( \cdot\right )$ is an indicator function, $\tau$ is a temperature hyper-parameter. In practical implementation, $g\left ( \cdot \right )$ is an MLP with one hidden layer followed by a ReLU function. The intra-CL can be written as:
\begin{equation}
\label{equation7}
    L_{intra}=\frac{1}{N\times N_{qry}}\sum_{j=1}^{N}\sum_{i=1}^{N_{qry}}l_{intra}\left ( x_i^{qry}, c'_j\right )
\end{equation}

Inter-CL contrasts query samples against the head instances from the imbalanced learning branch to distinguish the head- and tail features. For inter-CL, the positive samples are the same as intra-CL, for simplicity of notation, we still leverage $c'_j$ to stand for them. But negative samples are the head instances of the imbalanced learning branch.

Let $x_i^h$ represent a head instance and we have $N_{head}$ such instances. Similarly, inter-CL is defined as:
\begin{equation}
\label{equation8}
    s'_{i,h}=\frac{g\left( f_{\theta} \left( x_{i}^{qry}\right)\right)\cdot g\left( f_{\theta} \left( x_{i}^{h}\right)\right)}{\left \| g\left( f_{\theta} \left( x_{i}^{qry}\right)\right)\right \|\times \left \| g\left( f_{\theta} \left( x_{i}^{h}\right)\right)\right \|}
\end{equation}
\begin{equation}
\label{equation9}
    l_{inter}\left ( x_{i}^{qry},c'_{j}\right )=-\log\frac{\exp\left ( s_{i,j}/\tau\right )}{\sum_{h=1}^{N_{head}}\exp\left ( s'_{i,h}/\tau\right )}
\end{equation}
\begin{equation}
\label{equation10}
    L_{inter}=\frac{1}{N\times N_{qry}}\sum_{j=1}^{N}\sum_{i=1}^{N_{qry}}l_{inter}\left ( x_{i}^{qry},c'_{j}\right )
\end{equation}
where $s'_{i,h}$ is the cosine similarity of $x_i^{qry}$ and $x_i^h$, and $s_{i,j}$ is that of $x_i^{qry}$ and $c'_j$. Finally, the total loss of CoLB is:
\begin{equation}
\label{equation11}
    L_{con}=L_m+L_{intra}+L_{inter}\times \lambda
\end{equation}
where $\lambda$ is a hyper-parameter that controls the loss contribution of inter-CL. For the metric loss and intra-CL, they play their part inside the CoLB, and we equally set their weights as 1.

\subsection{The objective function}
With the above two branch losses, we finally define our training objective as:
\begin{equation}
\label{equation12}
    L=\alpha \times L_{imb}+\left ( 1-\alpha\right )\times L_{con}
\end{equation}
where $\alpha$ controls the importance of the two branches during the training phase. The larger $\alpha$ is, the more attention the model pays to the imbalanced learning branch. Since the imbalanced learning branch plays the most critical role, we focus on it first, and then gradually turn the focus towards CoLB. To meet this expectation, $\alpha$ should decrease as the training process progresses. So the parabolic decay function is chosen among all the elementary functions \citep{zhou2020bbn}. Let $T_{max}$ denote the total training epoch, and $T$ denote the current training epoch, $\alpha$ is defined as $\alpha =1-\left ( T/T_{max}\right )^2$. The training process of DB-LTR is provided in Algorithm \ref{algorithm1}.
\begin{algorithm}
\renewcommand{\algorithmicrequire}{\textbf{Input:}} 
\renewcommand{\algorithmicensure}{\textbf{Output:}}
    \caption{The training process of DB-LTR.}
    \label{algorithm1}
    \begin{algorithmic}[1]
        \REQUIRE training set $\mathbb{D}$, the number of epochs $T_{max}$.
        \ENSURE feature backbone $f_{\theta}$, classifier $f_{\varphi}$, projection head $g\left( \cdot \right)$.
        \FOR{$T=1$ to $T_{max}$}
        \STATE Sample for imbalanced learning branch using the uniform sampler
        \STATE Sample for contrastive learning branch using the tail sampler
        \STATE Extract features using the feature backbone
        \STATE Predict the features of imbalanced learning branch according to Equation \ref{equation1}
        \STATE Compute the $L_{imb}$ according to Equation \ref{equation2}
        \STATE Calculate prototypes $c_{j}$ of the support set
        \STATE Predict the features of the query set according to Equation \ref{equation3}
        \STATE Compute the $L_{m}$ according to Equation \ref{equation4}
        \STATE Feed the extracted features into $g\left( \cdot \right)$
        \STATE Calculate prototypes $c'_{j}$ using the features that are mapped by $g\left( \cdot \right)$
        \STATE Calculate the cosine similarity $s_{i,j}$ according to Equation \ref{equation5}
        \STATE Compute the $L_{intra}$ according to Equation \ref{equation7}
        \STATE Calculate the cosine similarity $s'_{i,h}$ according to Equation \ref{equation8}
        \STATE Compute the $L_{inter}$ according to Equation \ref{equation10}
        \STATE Compute the $L_{con}$ according to Equation \ref{equation11}
        \STATE $\alpha =1-\left( \frac{T}{T_{max}}\right)^{2}$
        \STATE Compute the training objective $L$ according Equation \ref{equation12}
        \STATE Update $f_{\theta}$, $f_{\varphi}$ and $g\left( \cdot \right)$ using $L$
        \ENDFOR
    \end{algorithmic}
\end{algorithm}
\section{Experiments}
% \label{experiments}
\subsection{Experimental settings}
\label{eperimental_settings}
\textbf{Datasets.} We evaluate our method by performing experiments on three long-tailed datasets: CIFAR100-LT \citep{cui2019class}, ImageNet-LT \citep{liu2019large} and Places-LT \citep{liu2019large}. For CIFAR100-LT, the training set contains 100 categories, and the number of training samples per category is $n=n_i/\mu^{i/100}$, where $i$ is the class index, $n_i$ is the number of original training samples in class $i$, $\mu$ is an imbalance factor. In CIFAR100-LT, each image is of size $32\times 32$. For ImageNet-LT, the training set is sampled from the original ImageNet-2012 \citep{deng2009imagenet} following the Pareto distribution with the power value of 6, while the validation set and test set remain unchanged. There are 115.8K images with size $224\times 224$ from 1000 categories in the training set, in which the number of samples per category is between 5-1280, so the imbalance factor is 256. For Places-LT, it is constructed similarly to ImageNet-LT, containing 62.5K images from 365 categories with a maximum of 4980 images and a minimum of 5 images, i.e., a imbalance factor of 996. This dataset contains all kinds of scenes and the image size is $256\times 256$.

\textbf{Evaluation metrics.} We report the average top-1 classification accuracy (\%) and standard deviation over 5 different runs to evaluate our model over all classes. And we further create three splits of datasets following \citep{liu2019large} and report accuracy over them. They are Many-shot (more than 100 images), Medium-shot (20$\sim$100 images) and Few-shot (less than 20 images), respectively.

\textbf{Implementation.} For fair comparisons, we follow the network setting in \citep{cui2019class, liu2019large} and use ResNet-32 \citep{he2016deep} for CIFAR100-LT, ResNet-10 \citep{he2016deep} for ImageNet-LT and ResNet-152 pre-trained on the full ImageNet dataset \citep{kang2020decoupling} for Places-LT as the model network, respectively. We train the network with 200 epochs on CIFAR100-LT and warm up the learning rate to 0.1 in the first 5 epochs and decay it at the 160th epoch and 180th epoch by 0.1. For ImageNet-LT, the network is optimized with 90 epochs. Likewise, a learning rate warming up to 0.1 in the first 5 epochs is used and decayed by 0.1 at the 30th epoch and 60th epoch. 30 epochs and an initial learning rate of 0.01 with cosine learning rate schedule \citep{loshchilov2016sgdr} are used to update the model on Places-LT.

The models are trained by SGD algorithm with fixed momentum of 0.9 and weight decay of 0.0005 for all datasets. In all experiments, $\tau$ and $\lambda$ are set to 0.6 and 0.3, respectively. Unless otherwise stated, the batch size is set to 128 for the imbalanced learning branch. For CoLB, we empirically find that the tail sampler sampling data on the Medium- and Few-shot subsets poses the best performance. For $N$ and $N_{sup}$, they are set to 5 and 4 by default, respectively. The ablative analysis and experimental results are shown in Section \ref{ablation_study}. Therefore, the tail sampler randomly samples 5 classes and 5 samples (4 support samples and 1 query sample) for each class in a mini-batch.

\textbf{Comparison methods.} We compare DB-LTR with six categories of methods: (a)CE: A traditional deep CNN classification model. (b)Re-weighting: Focal Loss \citep{lin2017focal}, CB Loss \citep{cui2019class}, CFL \citep{smith2022cyclical}, LDAM-DRW \citep{cao2019learning}. (c)Data augmentation: Remix-DRW \citep{chou2020remix}, Bag of Tricks \citep{zhang2021bag}. (d)Calibration. LA \citep{menon2021long}, CDA-LS \citep{islam2021class}, LADE \citep{hong2021disentangling}. (e)Decouple method: cRT\citep{kang2020decoupling}, LWS\citep{kang2020decoupling}, DaLS \citep{wang2022dynamic}, MiSLAS\citep{zhong2021improving}. (f) Others: OLTR \citep{liu2019large}, BBN \citep{zhou2020bbn}, Hybrid-PSC \citep{wang2021contrastive}, TSC \citep{li2022targeted}.

% \usepackage{multirow}
% \usepackage[normalem]{ulem}
\begin{table*}[!h] \footnotesize
\centering
\caption{Top-1 accuracy (\%) on CIFAR100-LT with ResNet-32. $\dagger$ denotes results are our reproduction with released code. We report the mean and standard deviation of our method over five different runs.}\label{tab:table1}
\begin{tabularx}{\textwidth}{c|c|ccc}\hline
\multirow{2}{*}{Method}  & \multirow{2}{*}{MFLOPs} & \multicolumn{3}{c}{Imbalance factor}  \\ 
        &      & 100   & 50    & 10 \\    \hline
CE   &  69.76  & 38.32 & 43.85 & 55.71 \\ \hline
Focal Loss\citep{lin2017focal} &69.76   & 38.41 & 44.32 & 55.78  \\ 
CB Loss\citep{cui2019class}  &69.76    & 39.60 & 45.32 & 57.99  \\ 
CFL\citep{smith2022cyclical} $\dagger$ &69.76 & 42.71 & 48.66 & 60.87  \\ 
LDAM-DRW\citep{cao2019learning} &69.76 & 42.04 & 47.30 & 58.71  \\         \hline
Remix-DRW\cite{chou2020remix} &69.76   & 46.77  & -    & 61.23  \\ 
Bag of Tricks\citep{zhang2021bag} &69.76 & {\ul 47.73}  & 51.69   & -      \\  \hline
CDA-LS\citep{islam2021class} $\dagger$ &69.76    & 41.42     & 46.22     & 59.86         \\ 
LA\citep{menon2021long}  &69.76   & 43.89     & -      & -         \\ 
LADE\citep{hong2021disentangling}   &74.22   & 45.4     & 50.5     & 61.7          \\ \hline
cRT\citep{kang2020decoupling} & 69.76 & 43.30 & 47.37 & 57.86 \\
LWS\citep{kang2020decoupling} & 69.76 & 42.97 & 47.40 & 58.08 \\
DaLS\citep{wang2022dynamic} &69.76  & 47.68   & 51.90   & 61.34   \\  
MiSLAS\citep{zhong2021improving} & 69.76 & 47.0  & {\ul 52.3}  & {\ul 63.2}   \\   \hline
OLTR\citep{liu2019large} &71.2  & 41.4     & 48.1     & 58.3      \\ 
BBN\citep{zhou2020bbn}     &74.22    & 42.56       & 47.02       & 59.12       \\ 
Hybrid-PSC\citep{wang2021contrastive} &69.76  & 44.97      & 48.93       & 62.37         \\ 
TSC\citep{li2022targeted} & -  & 43.8      & 47.4     & 59.0       \\ \hline
DB-LTR (ours) &69.76 & \textbf{48.83\tiny $\pm $0.06} & \textbf{53.67\tiny $\pm $0.12} & \textbf{63.89\tiny $\pm $0.10} \\ \hline
\end{tabularx}
\end{table*}

\subsection{Experimental results}
\label{experimental_results}
\textbf{Experimental results on CIFAR100-LT.} \cref{tab:table1} displays the overall top-1 accuracy of different methods on CIFAR100-LT with different imbalance factors of 100, 50 and 10. It can be observed that DB-LTR delivers the best results in all situations. These results validate the effectiveness of our method. We note that the performance gap between DB-LTR and Hybrid-PSC \citep{wang2021contrastive} declines as the extent of data imbalance decreases. This result is mainly attributed to the fact that the model poses poorer adaptability to tail classes when imbalanced problem is more severe, and our CoLB can make the model more adaptive and bring more performance gains.

\textbf{Experimental results on ImageNet-LT.} \cref{tab:table2} presents the detailed results of our proposal and other methods on ImageNet-LT. Our DB-LTR still consistently achieves the best performance on all occasions. From the table we can see, BBN \citep{zhou2020bbn} improves the classification on Medium- and Few-shot subsets, while sacrificing negligible accuracy on Many-shot classes. This verifies the effectiveness of reversed sampling strategy, which is used to impel the decision boundary to be away from tail classes. Different from BBN that utilizes cross-entropy loss to train the feature backbone, we introduce inter-CL and intra-CL to learn better feature representation. Therefore, DB-LTR outperforms BBN on all occasions. Note that the overall accuracy of Bag of Tricks \citep{zhang2021bag} is exceedingly close to that of our DB-LTR. Such merit of Bag of Tricks may account for the scientific combination of many efficacious training tricks. However, its training procedure is intractable due to the cumbersome combination of tricks. In comparison, DB-LTR is simple and easy to implement, enabling us to tackle long-tailed recognition effectively.

% \usepackage[normalem]{ulem}
\begin{table*}[!h] \footnotesize
\setlength{\tabcolsep}{1pt}
\centering
\caption{Top-1 accuracy (\%) on ImageNet-LT with ResNet-10. $\dagger$ denotes results are our reproduction with released code. We report the mean and standard deviation of our method over five different runs.}
\label{tab:table2}
\begin{tabularx}{\textwidth}{cccccc}\hline
Method  & GFLOPs  & Many      & Medium         & Few        & Overall       \\ \hline
CE  & 0.89  & {\ul 49.4}           & 13.7         & 2.4          & 23.9       \\ \hline
Focal Loss\citep{lin2017focal} & 0.89   & 36.4         & 29.9         & 16.0         & 30.5         \\
CB Loss\citep{cui2019class}  & 0.89     & 43.1           & 32.9           & {\ul 24.0}     & 35.8           \\
CFL\citep{smith2022cyclical} $\dagger$ & 0.89  &  46.73  & 23.18  & 13.51  & 32.46 \\
LDAM-DRW\citep{cao2019learning} & 0.89     & 45.3         & 34.1       & 19.3       & 36.3      \\ \hline
CDA-LS\citep{islam2021class}  & 0.89      & -              & -            & -            & 35.68          \\ 
LA \citep{menon2021long} $\dagger$ & 0.89 &  51.64  &  38.96  &  22.63   &  41.54   \\   \hline
Bag of Tricks\citep{zhang2021bag}  & 0.89     & -        & -            & -          & {\ul 43.13} \\ \hline
cRT\citep{kang2020decoupling}    & 0.89       & -          & -           & -           & 41.8           \\
LWS\citep{kang2020decoupling}     & 0.89    & -          & -          & -           & 41.4           \\
DaLS\citep{wang2022dynamic}     & -     & -          & -           & -           & 42.43    \\ \hline
OLTR\citep{liu2019large}    & 0.91      & 43.2           & 35.1         & 18.5           & 35.6         \\ 
BBN\citep{zhou2020bbn}     & -      & 49.1           & {\ul 37.1}     & 20.4     & 37.7        \\  \hline
DB-LTR (ours) &0.89 & \textbf{53.44\tiny $\pm $0.04} & \textbf{39.59\tiny $\pm $0.03} & \textbf{27.60\tiny $\pm $0.07} & \textbf{43.28\tiny $\pm $0.03} \\ \hline
\end{tabularx}
\end{table*}

\textbf{Experimental results on Places-LT.} To further verify the effectiveness of our method, we conduct comparative experiments on the large-scale scene dataset, i.e., Places-LT, and report the results of Many-shot, Medium-shot and Few-shot subsets. The experimental results are shown in \cref{tab:table3}. The imbalanced learning branch of the DB-LTR deploys LDAM-DRW \citep{cao2019learning} algorithm, whose core idea is to address the data imbalance and to mitigate the bias towards the dominant majority classes. Nevertheless, the insufficiency of feature representation for tail classes has not been considered and addressed. The proposed DB-LTR tackles these problems simultaneously. Specifically, in CoLB, we recurrently train the model with tail data to enhance the adaptability of the model. Besides, two contrastive losses are calculated to reduce the intra-class variances and to distinguish head- and tail classes. Therefore, our method surpasses LDAM-DRW by 16.40\%, and the well-learned representation boost the overall performance. LWS \citep{kang2020decoupling} shows the best performance on Medium- and Few-shot subsets as it is equipped with class-balanced sampling to adjust the tail classifier with small weight norm. However, LWS under-learns head classes and achieves unsatisfactory results on the Many-shot classes. In comparison, our method improves the classification of tail classes and the feature representation concurrently, so it is still superior to LWS on the Many-shot and overall accuracy.

% \usepackage[normalem]{ulem}
\begin{table}[!h] \footnotesize
\setlength{\tabcolsep}{3pt}
\centering
\caption{Top-1 accuracy (\%) on Places-LT with ResNet-152. $\ddagger$ and $\dagger$ denote the results are borrowed from \citep{liu2019large} and reproduced by us with released code, respectively. We report the mean and standard deviation of our method over five different runs.}\label{tab:table3}
\begin{tabularx}{\textwidth}{cccccc}  \hline
Method   & GFLOPs     & Many       & Medium      & Few           & Overall        \\ \hline
CE  & 11.55   & {\ul 45.9}    & 22.4          & 0.36          & 27.2           \\ \hline
Focal Loss\citep{lin2017focal} $\ddagger$  & 11.55  & 41.1       & 34.8      & 22.4      & 34.6        \\
CB Loss\citep{cui2019class} $\dagger$   & 11.55    & 43.44      & 33.71      & 13.37      & 32.94          \\
CFL\citep{smith2022cyclical} $\dagger$   & 11.55     & 45.07      & 25.76     & 8.50    & 29.11     \\
LDAM-DRW\citep{cao2019learning} $\dagger$  & 11.55   & 43.25   & 31.76   & 16.68     & 32.74     \\ \hline
CDA-LS \citep{islam2021class} $\dagger$ & 11.55 &  42.34   &  33.13   &  18.42  &  33.61   \\ 
LA \citep{menon2021long} $\dagger$  & 11.55& 43.10 & 37.43 & 22.40& 36.31\\ \hline
cRT\citep{kang2020decoupling}   & 11.55        & 42.0      & 37.6       & 24.9       & 36.7           \\
LWS\citep{kang2020decoupling}   & 11.55     & 40.6     & \textbf{39.1} & \textbf{28.6} & 37.6      \\\hline
OLTR\citep{liu2019large}   & 11.56       & 44.7      & 37.0      & 25.3      & 35.9         \\  
BBN\citep{zhou2020bbn}    & -       & \textbf{49.1} & 37.1          & 20.4          & {\ul 37.7}     \\   \hline
DB-LTR (ours) &11.55 & 43.85\tiny $\pm $0.04 & {\ul 38.65}\tiny $\pm $0.06 & {\ul 26.93}\tiny $\pm $0.04 & \textbf{38.11\tiny $\pm $0.02} \\ \hline
\end{tabularx}
\end{table}

As shown in \cref{tab:table1,tab:table2,tab:table3}, the FLOPs of our method are close to those of previous methods. The additional computational overhead of DB-LTR mainly lies in the second branch, where DB-LTR applies a projection head $g\left( \cdot \right)$ to calculate the inter-CL and intra-CL. However, compared to the entire network structure, the computational cost of the projection head can be ignored.

\subsection{Ablation study}
\label{ablation_study}
\textbf{CoLB plugs in previous methods.} As mentioned before, our CoLB can be treated as a plug-and-play module and seamlessly combined with previous methods to further improve their performance. Experiments on CIFAR100-LT with imbalance factor of 100 for validating this are shown in \cref{tab:table4}. As shown in this table, CoLB improves the recognition performance of previous methods. Since CoLB learns a prototype \citep{snell2017prototypical} for each tail class and calculates a prototype-based metric loss, the model is more adaptive to tail classes, which results in noticeable performance improvement of the minority. On the other hand, the proposed CoLB selects positive pairs from tail classes, aligning them and repulsing head instances in the embedding space, to learn a discriminative feature space that possesses the property of intra-class compactness and inter-class separability. Hence, the distinguishability of the decision boundary gets upgraded. The approach MiSLAS \citep{zhong2021improving} utilizes mixup \citep{zhang2017mixup} to find better representation, when combined with CoLB, it still gets 1.06 points performance gains.

% \usepackage{multirow}
\begin{table}[!h]
\centering
\caption{Evaluation of CoLB plugging in previous methods. $\dagger$ denotes results are reproduced with released code. The green numbers in round brackets stand for the performance gains of leading CoLB into other methods. $\surd$ and $\times$ indicate CoLB has / has not been incorporated into the corresponding methods.}\label{tab:table4}
\begin{tabular}{c|c|c}  \hline
Method                        & CoLB & Top-1 accuracy (\%)                  \\ \hline
                              & $\times$    & 38.32                                \\
\multirow{-2}{*}{CE} & $\surd$    & 43.65 (\textcolor{green}{+5.33}) \\ \hline
                              & $\times$    & 38.41          \\
\multirow{-2}{*}{Focal Loss\citep{lin2017focal}}  & $\surd$    & 43.14 (\textcolor{green}{+4.73})  \\ \hline
                              & $\times$    & 42.71$\dagger$           \\
\multirow{-2}{*}{CFL\citep{smith2022cyclical}}         & $\surd$    & 45.16 (\textcolor{green}{+2.45})   \\ \hline
                              & $\times$    & 46.52$\dagger$           \\
\multirow{-2}{*}{Remix-DRW\citep{chou2020remix}}   & $\surd$    & 48.46 (\textcolor{green}{+1.94})   \\ \hline
                              & $\times$    & 41.42$\dagger$            \\
\multirow{-2}{*}{CDA-LS\citep{islam2021class}}      & $\surd$    & 46.98 (\textcolor{green}{+5.56})   \\ \hline
                              & $\times$    & 47.0            \\
\multirow{-2}{*}{MiSLAS\citep{zhong2021improving}}      & $\surd$    & 48.06 (\textcolor{green}{+1.06})   \\ \hline
\end{tabular}
\end{table}

\textbf{CoLB with different losses.} Intending to investigate the effect of the three losses in CoLB, i.e., inter-CL, intra-CL and metric loss, we carry on experiments on CIFAR100-LT with imbalance factor of 100 and ImageNet-LT. The results are shown in \cref{tab:table5}. The first line of the table shows the result of LDAM-DRW \citep{cao2019learning}, i.e, only using the imbalanced learning branch in DB-LTR.

It can be observed that the metric loss respectively delivers 10.15\% and 10.60\% performance gains on CIFAR100-LT and ImageNet-LT compared with LDAM-DRW, which is attributed to the better adaptation to tail classes. Based on the metric loss, the performance obtains further improvement when integrating intra-CL or inter-CL into CoLB. It is noteworthy that compared with the combination of metric loss and intra-CL, that of  metric loss and inter-CL achieves better recognition performance. Due to the imbalance of class label and the scarcity of tail samples, it is difficult for the model to depict the real feature distribution of tail classes, resulting in mixed head- and tail features. Inter-CL can separate head- and tail classes from each other, while intra-CL makes the class boundaries of minority classes distinguishable. Hence, when cooperating with metric loss, the accuracy gains brought by inter-CL are greater than intra-CL. The combination of inter-CL and intra-CL is also significantly better than LDAM-DRW, corroborating the better feature representation is conducive to model performance. When the metric loss is further added to the combination of inter-CL and intra-CL, the model achieves the best performance.

\begin{table}[!h] \footnotesize
\setlength{\tabcolsep}{6pt}
\centering
\caption{Evaluation of CoLB with different losses. $\surd$ denotes using the corresponding loss in experiments. Top-1 accuracy (\%) is reported on CIFAR100-LT with an imbalance factor of 100 and ImageNet-LT.}\label{tab:table5}
\begin{tabularx}{\textwidth}{
>{\centering\arraybackslash}X
>{\centering\arraybackslash}X
>{\centering\arraybackslash}X
>{\centering\arraybackslash}X
>{\centering\arraybackslash}X
}   \hline
Metric loss & Intra-CL & Inter-CL & Accuracy on CIFAR100-LT & Accuracy on ImageNet-LT \\ \hline
         &          &       & 42.04    & 36.3\\ 
$\surd$  &          &       & 46.31    & 40.15\\
$\surd$  & $\surd$  &       & 47.48    & 41.48\\
$\surd$  &          & $\surd$  & 47.61 & 41.67\\
         & $\surd$  & $\surd$  & 48.21 & 42.29\\
$\surd$  & $\surd$  & $\surd$  & \textbf{48.88} & \textbf{43.23}   \\ \hline
\end{tabularx}
\end{table}

\begin{figure*}[!b]
    \centering
    \includegraphics[width=0.95\textwidth,height=0.45\textwidth]{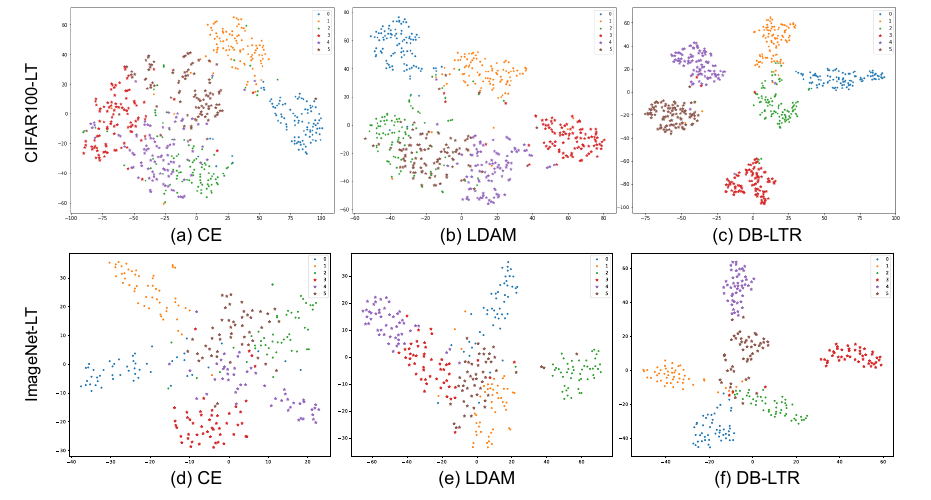}
    \caption{The t-SNE visualization of embedding space obtained using CE, LDAM and DB-LTR. The first and the second rows visualize results on the CIFAR100-LT test set and the ImageNet-LT test set, respectively. The dots of No. 0$\sim$2 denote the features of head classes and the stars of No. 3$\sim$5 stand for tail features. Zoom in for details.}\label{fig:feature}
\end{figure*}

\textbf{Visualization of feature distribution.} In order to verify that DB-LTR can learn a better feature space with intra-class compactness and inter-class separability, we visualize the learned feature distributions of our method and other approaches. The compared methods include CE, LDAM \citep{cao2019learning} and DB-LTR. We use t-SNE \citep{van2008visualizing} visualization technology, a common method to detect feature quality in visual images, to display the comparison results in Figure \ref{fig:feature}. From the figure we can see, the quality of the features learned by CE is the worst, in which most of the features for different categories mix together and the decision boundaries are ambiguous. It implies that cross-entropy loss is not suitable for feature learning in the long-tailed scenarios because of the extreme class imbalance and the lack of tail samples. LDAM encourages model to learn different margins for different classes, which improves the decision boundaries to some extent. However, the intra-class features trained by LDAM are scattered in the embedding space and the issue of inadequate feature learning for tail classes has not been solved satisfactorily. In contrast with CE and LDAM, our DB-LTR learns high-quality features and decision boundaries. As Figure \ref{fig:feature} (c) and (f) illustrate, the features within class closely gather together, revealing that the intra-class variances are quite small. Meanwhile, there are more obvious decision boundaries. The results prove the effectiveness and practicality of CoLB in assisting the imbalanced learning branch to learn compact and separable feature space and discriminative decision boundary.

\begin{figure*}[!h]
    \centering
    \includegraphics[width=0.95\textwidth,height=0.4\textwidth]{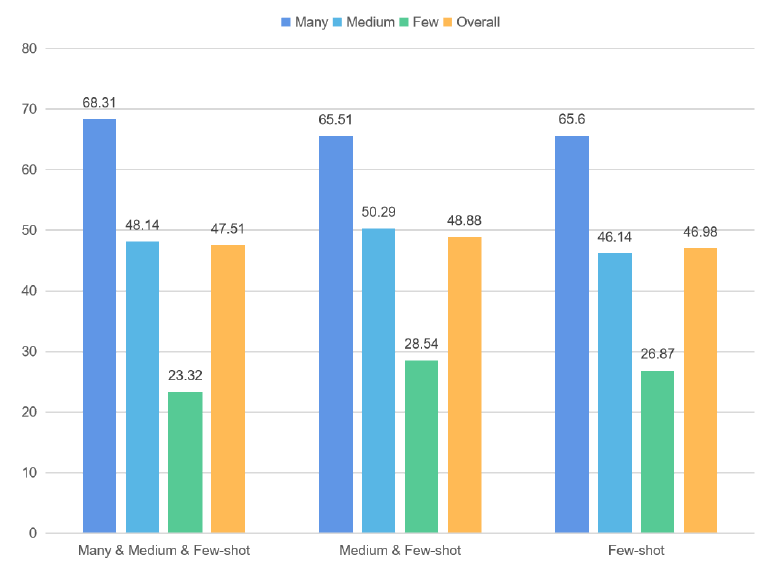}
    \caption{The detailed results of the tail sampler sampling data on the whole dataset, on the Medium- and Few-shot subsets, and on the Few-shot subset, respectively.}\label{fig:sampling}
\end{figure*}

\textbf{Sampling strategy of tail sampler.} In Figure \ref{fig:sampling}, we discuss the impact of the sampling strategy for the tail sampler on the model performance. We conduct experiments on CIFAR100-LT with imbalance factor of 100. As shown in Figure \ref{fig:sampling}, when the training data of CoLB is sampled on Medium- and Few-shot subsets, the model achieves the best performance due to the more balanced learning between head- and tail classes. When the tail sampler samples data on the whole dataset, the model shows inferior overall accuracy. The main reason is that the training data of CoLB is dominated by head classes, which brings more performance improvement to the majority yet less one to the minority. When the data of few-shot subset is sampled by the tail sampler, the overall accuracy is the worst, because Medium-shot classes receive less focus and pose the lowest accuracy. In a nutshell, tail sampler sampling on Medium- and Few-shot classes better balances the learning law of the model, thus delivering the best performance.

\textbf{Effect of $N$ and $N_{sup}$.} The proposed CoLB supervises the model with $N$ categories and $N_{sup}$ support samples for each category. To investigate the sensitivity of our model to $N$ and $N_{sup}$, we conduct experiments with different values of $N$ and $N_{sup}$. From Table \ref{tab:table6} we can see, as the value of $N_{sup}$ increases, the performance of our model initially improves, however after a certain point it starts to drop. It is attributed to the fact that the larger $N_{sup}$ is, the more high-quality the learned prototype is, the better results the model poses. However, when $N_{sup}$ increases to 10, the computed prototype can nearly impeccably represent the true prototype of the corresponding class. Therefore, we draw a conclusion that our method is susceptible to the number of support samples. Despite that, since there are extremely few samples for some tail categories when the imbalance factor becomes large, e.g., imbalance factor of 256 for ImageNet-LT, we set $N_{sup}$ to 4 by default. Table \ref{tab:table7} demonstrates a large $N$ will deteriorate the model performance because of the overfitting problem, which is similar to oversampling tail classes \citep{chawla2002smote, more2016survey}. In a nutshell, we carry on experiments under 5-way 4-shot setting.

\begin{table}[!h]
    \begin{minipage}{0.45\linewidth}
        \centering
        \caption{The ablation study of the hyper-parameter $N_{sup}$. The results are reported on CIFAR100-LT with an imbalance factor of 10 and $N=5$ (5-way $N_{sup}$-shot).}\label{tab:table6}
        \resizebox{0.9\textwidth}{!}{
        \begin{tabular}{c|c}    \hline
        $N_{sup}$    &  Top-1 accuracy (\%)    \\  \hline
        1     &   62.93    \\ \hline
        4  & 63.89 \\ \hline
        10     &   64.14 \\ \hline
        20     &   62.73  \\ \hline
        \end{tabular}
        }
    \end{minipage}
    \hfill
    \begin{minipage}{0.45\linewidth}
        \centering
        \caption{The ablation study of the hyper-parameter $N$. The results are reported on CIFAR100-LT with an imbalance factor of 100 and $N_{sup}=4$ ($N$-way 4-shot).}\label{tab:table7}
        \resizebox{0.9\textwidth}{!}{
        \begin{tabular}{c|c}   \hline
        $N$ &  Top-1 accuracy (\%)   \\ \hline
        5 & 48.83 \\    \hline
        10 & 48.27   \\ \hline
       \end{tabular}
        }
    \end{minipage}
\end{table}

\begin{figure}
    \centering
    \includegraphics[width=0.95\textwidth,height=0.4\textwidth]{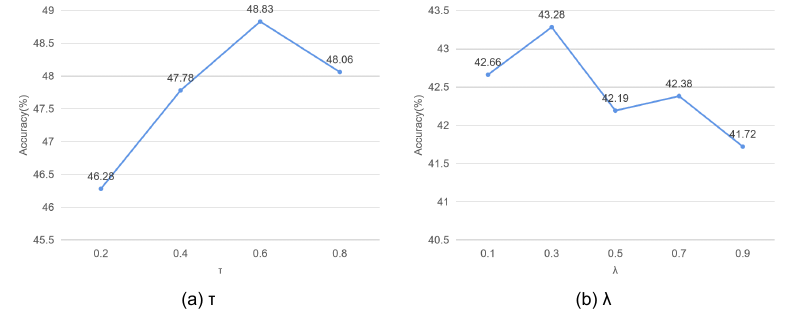}
    \caption{The influence of hyper-parameter $\tau$ and $\lambda$ on model performance. The results are produced on (a) CIFAR100-LT with an imbalance factor of 100 and (b) ImageNet-LT, respectively.}
    \label{fig:hyperpamrm}
\end{figure}

\textbf{The hyper-parameter $\tau$ and $\lambda$.} In Figure \ref{fig:hyperpamrm}, we explore the influence of $\tau$ and $\lambda$ on the performance of DB-LTR. As illustrated in the figure, $\tau=0.6$ makes the model perform the best. And setting $\lambda$ with a small or large value brings inferior performance gains due to the under and over regularization of inter-CL. Therefore, we set the values of $\tau$ and $\lambda$ to 0.6 and 0.3 in experiments.

\section{Conclusion}
\label{conclusion}
In this paper, we propose a Dual-Branch Long-Tailed Recognition (DB-LTR) model to deal with long-tailed problems. Our model consists of an imbalanced learning branch and a Contrastive Learning Branch (CoLB). The imbalanced learning branch can integrate common imbalanced learning methods to deal with the data imbalance issue. With the introduction of the prototypical network and contrastive learning, CoLB can learn well-trained feature representation, thus improving the adaptability of the model to tail classes and the discriminability of the decision boundaries. Specifically, the prototype-based metric loss improves the ability of the model to recognize tail classes, the inter-branch contrastive loss and the intra-branch contrastive loss make the learned feature space more compact and separable. CoLB is a plug-and-play module, when other imbalanced learning methods are combined with CoLB,  their recognition performance is improved. The proposed method outperforms the comparative methods on three common long-tailed datasets, proving its effectiveness and competitiveness.

\section*{Declaration of competing interest}
The authors declare that they have no known competing financial interests or personal relationships that could have appeared to influence the work reported in this paper.

\section*{Acknowledgements}
This work is supported by the National
Natural Science Foundation of China (No. 62176095), and
the Guangdong Provincial Key Laboratory of Artificial Intelligence in Medical Image Analysis and Application (No.2022B1212010011).








% \end{thebibliography}

% customed bibliography
\bibliographystyle{elsarticle-harv} 
\bibliography{harv}
\end{document}